\documentclass{article}
\usepackage{spconf,amsmath,graphicx}

\usepackage{booktabs}
\usepackage{amsfonts}

\usepackage{times}
\usepackage{soul}
\usepackage{url}
\usepackage[hidelinks]{hyperref}
\usepackage[utf8]{inputenc}
\usepackage[small]{caption}
\usepackage{graphicx}
\usepackage{amsmath}
\usepackage{amsthm}
\usepackage{booktabs}
\usepackage{algorithm}
\usepackage{algorithmic}
\usepackage[switch]{lineno}
\usepackage{enumitem}

\usepackage{graphicx}
\usepackage{algorithmic}
\usepackage{pifont}
\usepackage[most]{tcolorbox}
\usepackage{booktabs}
\usepackage{amsfonts}

\usepackage{hyperref}
\usepackage{url}
\usepackage{listings}
\usepackage{tipa}

\title{Character-Centered Dialogue Generation from Scene-Level Prompts}

\name{
    Taewon Kang and Ming C. Lin
}
\address {University of Maryland at College Park, United States \\
    {\large \texttt{taewon@umd.edu, lin@umd.edu}} \\
}

\begin{document}
\maketitle
\begin{abstract}
    Recent advances in scene-based video generation enable coherent visual narratives from structured prompts, yet a key aspect of storytelling—\textit{character-driven dialogue and speech}—remains underexplored. We present a modular pipeline that transforms action-level prompts into visually and auditorily grounded dialogue, enriching scene-based storytelling with natural voice and character expression. Our method takes a pair of prompts per scene, defining the setting and character behavior. While a story generation model such as Text2Story produces the visual scene, we focus on generating expressive, character-consistent utterances grounded in both the prompts and a representative scene image. A pretrained vision-language encoder extracts high-level visual semantics, which are combined with structured prompts to guide a large language model for dialogue synthesis. To maintain contextual and emotional consistency across scenes, we introduce a \textit{Recursive Narrative Bank}, a speaker-aware, temporally structured memory that accumulates each character’s dialogue history. Inspired by Script Theory, this design enables dialogue that reflects evolving goals, social context, and narrative roles. Finally, we render each utterance as expressive, character-conditioned speech, producing fully voiced, multimodal video narratives. Our training-free framework generalizes across diverse story settings, providing a scalable solution for coherent, character-grounded audiovisual storytelling.
\end{abstract}


\section{Introduction}

\begin{figure}[h]
\centering
\includegraphics[width=0.85\linewidth]{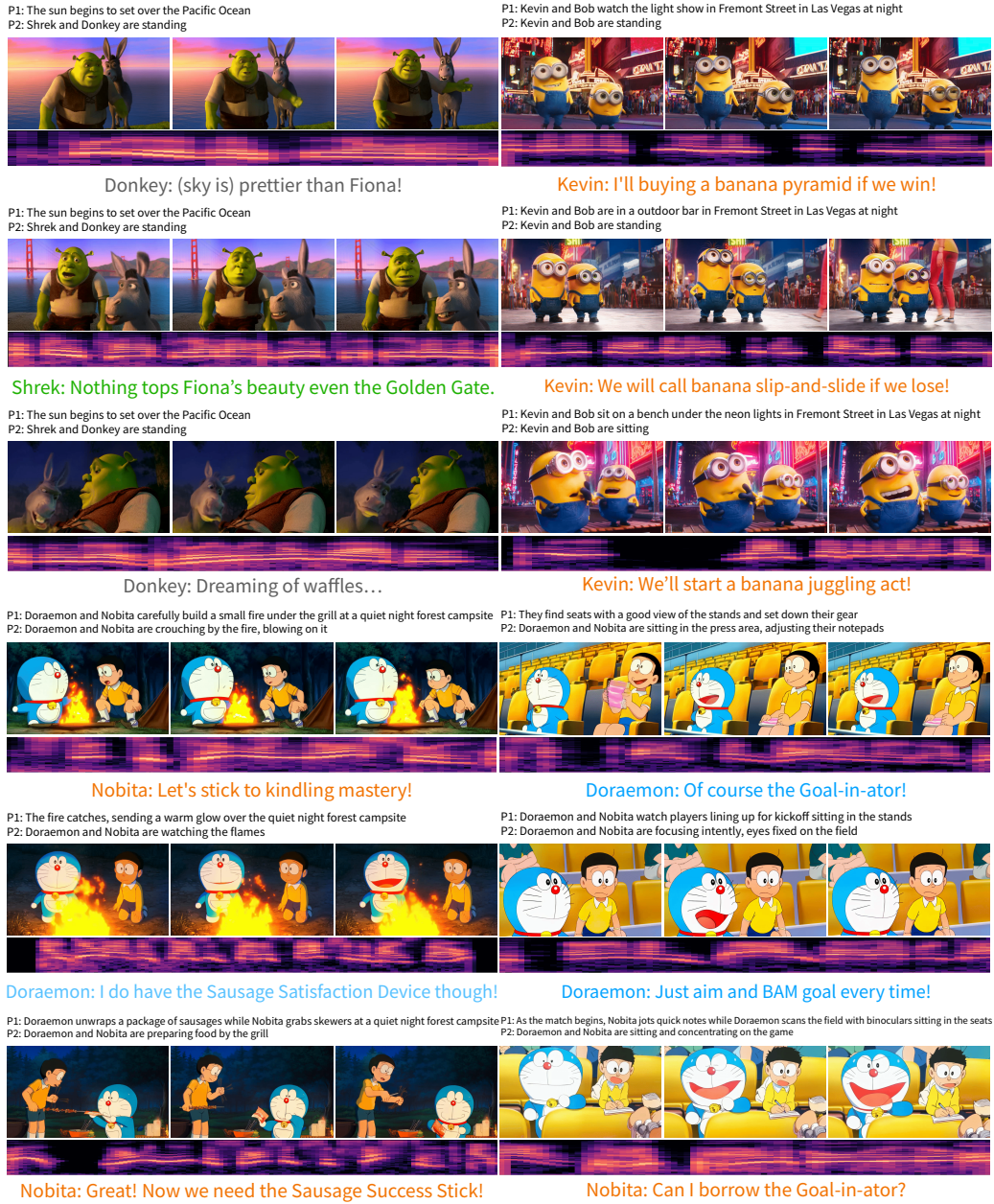}
\caption{\small \textbf{Storytelling across Thematic Settings.} We showcase multimodal storytelling across four distinct settings using animated characters. \textbf{(a) Urban Exploration in San Francisco} (Top, L) Shrek and Donkey reflect on the beauty of the Pacific sunset and the Golden Gate, blending humorous banter with visual splendor. \textbf{(b) Urban Exploration in Las Vegas} (Top, R): Kevin and Bob (Minions) navigate the vibrant nightlife of Fremont Street with playful bets, envisioning banana pyramids and juggling acts under neon lights. \textbf{(c) Outdoor Cooking Show} (Bottom, L): Doraemon and Nobita build a campfire and engage in a light-hearted cooking segment, featuring whimsical gadgets like the "Sausage Satisfaction Device." \textbf{(d) Sports Reporting} (Bottom, R): Doraemon and Nobita become animated sports commentators, highlighting game moments with imaginative tools, like the "Goal-in-ator." (Zoom in to see the figure)}
\label{fig:characters}
\end{figure}

Multimodal storytelling aims to generate coherent narratives by integrating visual, textual, and auditory modalities. Recent progress in storytelling video generation has enabled systems such as Text2Story~\cite{kang2025text2story} to synthesize temporally coherent scenes from structured prompts. However, these pipelines primarily emphasize visual consistency and largely overlook linguistic expressivity, particularly \textit{character-driven dialogue and speech}, which are essential for emotional depth and narrative realism.

A central challenge lies in transforming high-level scene and action prompts into dialogue that is natural, context-aware, and consistent with a character’s persona. While prompts such as \textit{“The sun begins to set over the Pacific Ocean”} and \textit{“Shrek and Donkey are standing”} can be visually grounded, generating appropriate dialogue requires awareness of prior exchanges and character relationships. As illustrated in Fig.~\ref{fig:characters}, dialogue that evolves across scenes can convey continuity, rivalry, and character dynamics—capabilities that captioning or template-based methods lack due to their inability to model dialogue history or evolving interactions. To address this gap, we present a pipeline that generates coherent, character-driven dialogue and speech from structured scene prompts. Each scene is described by a pair of prompts specifying the setting and character action. While visual content is produced by a story generation model such as Text2Story, our focus is on synthesizing expressive, persona-consistent dialogue grounded in both the prompts and the scene image. A pretrained vision-language encoder extracts high-level semantic features from a representative keyframe, which are combined with structured prompts to guide a large language model for dialogue generation.

To maintain contextual and emotional consistency across scenes, we introduce a \textit{Recursive Narrative Bank (RNB)}, a speaker-aware memory that accumulates and organizes dialogue over time. Unlike stateless GPT-style APIs or chat-based interfaces that retain unstructured history, RNB maintains \textbf{role-conditioned, temporally structured memory for each character}, enabling contextually appropriate and emotionally consistent dialogue generation across scenes. Conditioning the language model on this structured memory, together with current prompts and visual context, allows characters to naturally reference prior events and exhibit evolving emotions. This design is inspired by \textbf{Script Theory}~\cite{schank2013scripts,bower1979scripts,wilensky1983planning}, which models human behavior through structured sequences of roles and events, and operationalizes it within a training-free prompt design for long-form, multimodal dialogue generation.

We further enhance immersion by rendering generated dialogue as expressive, character-consistent speech using a reference-driven voice synthesis approach. Short audio clips with aligned transcripts implicitly capture character-specific vocal traits and prosody, producing emotionally grounded speech without additional model training. Our modular, training-free framework supports diverse characters, scenes, and storytelling styles, enabling coherent multimodal narratives with synchronized dialogue and voice. Through extensive quantitative and qualitative evaluation, including a human subject study, we demonstrate that our approach significantly improves dialogue quality, character consistency, and user preference, providing a scalable foundation for character-driven audiovisual storytelling.

\section{Related Work}
\subsection{Storytelling Video Generation}
Storytelling video generation has progressed from early script-to-video and image-sequence pipelines toward multimodal systems capable of producing longer visual narratives. These approaches primarily focus on generating temporally coherent scenes or video clips from structured prompts, establishing important foundations for visual storytelling. However, they are largely designed to model visual consistency and motion dynamics, and do not explicitly address character-driven dialogue, speaker identity, or expressive speech generation. Several recent systems leverage large language models to decompose scripts into scene-level instructions and guide video synthesis, enabling more structured long-form generation. While effective at organizing visual content, such methods treat dialogue—if present at all—as auxiliary narration rather than as a first-class, character-grounded signal. In particular, they lack mechanisms for modeling evolving character interactions, maintaining realized dialogue history, or synchronizing utterances with visual context and speech. Text2Story~\cite{kang2025text2story} provides a structured, training-free framework for long-form visual storytelling from prompt sequences, achieving improved visual coherence across scenes. Nevertheless, it remains limited to visual output and does not generate character dialogue or voice. Our work builds on the assumption that visual storytelling pipelines can provide scene-level grounding, and instead focuses on the complementary problem of generating expressive, character-consistent dialogue and speech conditioned on scene prompts and visual context. By augmenting existing visual generation systems with dialogue modeling and speech synthesis, our approach extends storytelling beyond visual coherence toward fully realized audiovisual narratives.

\vspace*{-1em}
\subsection{Multimodal Storytelling}
Recent multimodal storytelling systems integrate text, vision, and audio to improve narrative expressiveness. Methods such as Improving Visual Storytelling with Multimodal LLMs~\cite{lin2024improving} and SEED-Story~\cite{yang2024seed} leverage vision-language models for coherent image-text narratives, but do not support spoken dialogue or character-specific utterances. Other approaches incorporate audio through narration or background sound. LLaMS~\cite{zang2024let} and Art of Storytelling~\cite{arif2024art} enhance narrative richness with commonsense reasoning or co-created narration, yet produce narrator-style speech without character identity. Sound of Story~\cite{bae2023sound} focuses on ambient audio, while StoryAgent~\cite{sohn2024words}, Dialogue Director~\cite{zhang2024dialogue}, and MM-StoryAgent~\cite{xu2025mm} orchestrate multimodal assets or scripted narration without generating character-grounded dialogue or speech. A separate line of work addresses audio- or image-driven character animation, including EMO~\cite{tian2024emo}, Hallo3~\cite{cui2025hallo3}, Omnihuman-1~\cite{lin2025omnihuman}, and MoCha~\cite{wei2025mocha}. While effective for portrait animation, these methods generate new videos from a single image and speech input, do not support multi-character interaction, and are not designed for long-form, scene-based storytelling; MoCha further requires training and explicit emotional control. While many of these systems adopt GPT-based or multimodal architectures, \textbf{none jointly achieve character-grounded dialogue generation and expressive, synchronized speech within scene-based storytelling}. Our approach is not a direct application of a language model, but a purpose-built pipeline that grounds behavior prompts into utterances and renders character-consistent speech. \textbf{The observed performance gains arise from the integration of visual grounding, dialogue modeling, and speech synthesis, rather than the backbone model alone}, enabling expressive character-driven narratives beyond prior multimodal storytelling systems.

\subsection{Text-to-Video Diffusion, Dialogue Generation from Visual Context, and Text-to-Speech}
Text-to-video diffusion models have achieved substantial progress in temporal control and visual quality. However, models such as Mochi~\cite{mochi} and CogVideoX~\cite{yang2024cogvideox} still struggle with coherent multi-shot storytelling and fine-grained scene grounding, limiting their applicability to narrative dialogue generation. Dialogue generation from visual input has traditionally relied on full scripts, captions, or summaries, which are insufficient for modeling interaction or persona. Although vision-language models provide strong cross-modal representations, they are not designed for character-consistent dialogue generation across scenes. We adopt BLIP~\cite{li2022blip} for its generative compatibility and structured prompting. For speech synthesis, Tacotron~\cite{wang2017tacotron} and Bark~\cite{bark2023} generate fluent speech but lack character specificity, while SesameAILabs’ CSM~\cite{sesame2024csm1b} enables expressive, identity-aware synthesis through reference audio. Building on this capability, we employ a reference-driven approach that aligns generated dialogue with character identity and visual context, enabling expressive speech for scene-grounded storytelling.

\section{Method}
\label{sec:method}

\begin{figure*}[htb!]
\begin{center}
\vspace*{-7em}
\includegraphics[width=0.8\linewidth]{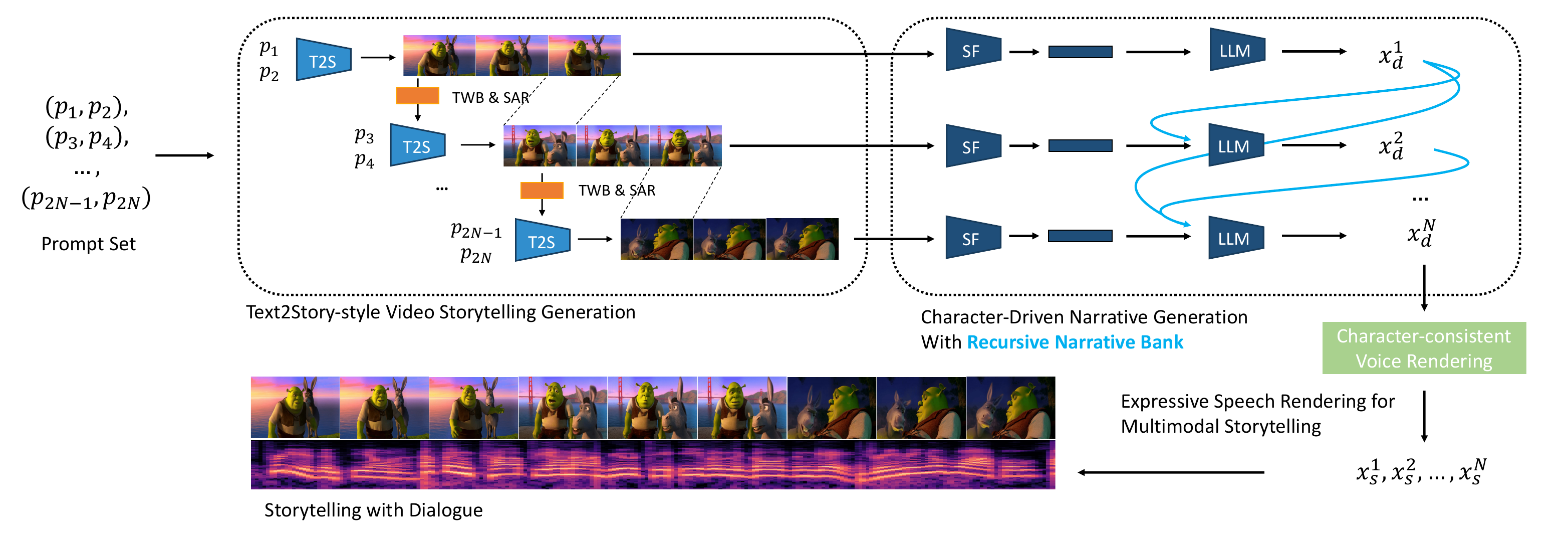}
\vspace*{-2em}
\end{center}
    \caption{\textbf{Overview of our proposed multimodal storytelling framework.} 
    Given a sequence of paired prompts \((p_1, p_2), (p_3, p_4), \dots, (p_{2N-1}, p_{2N})\), our system generates coherent video scenes, natural dialogue, and expressive speech. A story generation model (e.g., Text2Story (T2S)) synthesizes short video clips for each prompt pair, which are blended using Time-Weighted Blending (TWB) and refined with Semantic Action Representation (SAR) to ensure temporal coherence. For each scene, we extract a representative frame using \texttt{SceneFeat} (SF) to obtain a semantic visual feature. This, along with the current prompt pair and a Recursive Narrative Bank of prior dialogue, is used to prompt a large language model for generating character-consistent dialogue. Finally, the generated utterance is converted into speech with reference-driven conditioning to produce high-fidelity, emotionally expressive voice output. The result is a fully voiced, visually grounded, and narratively coherent story video.}
\label{method_figure}
\vspace*{-1em}
\end{figure*}

Our goal is to generate character-grounded dialogue and expressive speech from structured video scene prompts. Each scene is specified by a prompt pair: a setting prompt $p_1$ and a character action prompt $p_2$, following the formulation of Text2Story~\cite{kang2025text2story}. The overall process is defined as:
\vspace*{-1em}
\begin{equation}
(x_v, x_d, x_s) = \mathcal{F}(p_1, p_2, \mathcal{H}),
\end{equation}
\vspace*{-1em}

where $x_v$ denotes the generated visual scene, $x_d$ the dialogue, $x_s$ the synthesized speech, and $\mathcal{H}$ the dialogue history maintained by a \textit{Recursive Narrative Bank (RNB)}. While $\mathcal{F}$ is implemented as a three-stage system, we interpret it as a \textbf{factorized multimodal mapping} with an explicit intermediate narrative state:
\vspace*{-1em}
\[
\mathcal{F}
= \mathcal{F}_{\text{speech}}
\circ
\mathcal{F}_{\text{dialogue}}
\circ
\mathcal{F}_{\text{vision}}.
\]
\vspace*{-1em}

This interpretation preserves the original computation while making the role of narrative state propagation explicit.

\vspace*{-1em}
\subsection{Scene Visualization}
To obtain visual grounding, we assume a story generation model (e.g., Text2Story) that synthesizes a short video $x_v$ from $(p_1, p_2)$. We uniformly sample $K$ keyframes $\{I_1,\dots,I_K\}$ from $x_v$:
\begin{equation}
I_k = \texttt{SampleFrame}(x_v, t_k), \quad k \in [1, K],
\end{equation}
where $t_k$ are evenly spaced timestamps. For efficiency, we use the middle frame $I = I_{\lfloor K/2 \rfloor}$ as a representative scene image for dialogue generation.

\vspace*{-1em}
\subsection{Dialogue Generation}
Dialogue generation is conditioned on the current scene and accumulated narrative context. We define the scene prompt as $p = p_1 + \texttt{``. ''} + p_2$. Given the representative frame $I$, we extract a high-level visual representation using a pretrained vision-language encoder:
\begin{equation}
c = \texttt{SceneFeat}(I),
\end{equation}
where $\texttt{SceneFeat}$ is implemented using BLIP~\cite{li2022blip}. The resulting representation is decoded into a caption and used to provide visually grounded context to the language model.

To maintain cross-scene coherence, we introduce a \textbf{Recursive Narrative Bank (RNB)} $\mathcal{H}$ that stores prior dialogue. At scene step $t$:
\begin{equation}
\mathcal{H}_t = \{x_d^{(t-1)}, x_d^{(t-2)}, \dots, x_d^{(t-N)}\},
\end{equation}
where $x_d^{(i)}$ denotes dialogue generated at step $i$. In our setting, $N$ spans all previous scenes, enabling dialogue generation to reflect long-range narrative context. Importantly, we interpret $\mathcal{H}_t$ as a \emph{recursive narrative state} that evolves over scenes, rather than a flat token buffer. Formally, the state is updated by a deterministic, non-parametric transition:
\[
\mathcal{H}_t = \Phi(\mathcal{H}_{t-1}, x_d^{(t-1)}),
\]
which corresponds to appending the latest utterance and applying truncation when needed. Under this formulation, earlier dialogue affects generation only through the current state, i.e.,
\[
p(x_d^{(t)} \mid \{x_d^{(i)}\}_{i<t}, c, p) = p(x_d^{(t)} \mid \mathcal{H}_t, c, p).
\]

Dialogue is generated as:
\begin{equation}
x_d = \texttt{LLM}(c, p, \mathcal{H}_t),
\end{equation}
where the language model is conditioned via a structured prompt:
\begin{align}
\texttt{Input} = \texttt{NarrativePrompt}(c, p, \mathcal{H}_t) = \texttt{[Scene]} \, p \notag \\
\quad || \, \texttt{[Image]} \, c \, || \, \texttt{[DialogueMemory]} \, \mathcal{H}_t .
\end{align}
\vspace*{-2em}

This explicit decomposition separates scene description, visual grounding, and narrative state, enabling contextually appropriate and persona-consistent utterances. Additional details are provided in Appendix~\ref{supp:RNB}.

\subsection{Speech Synthesis}
Given dialogue $x_d$, we synthesize speech $x_s$ using a reference-driven voice synthesis approach. Instead of predefined speaker embeddings or fine-tuning, speech generation is conditioned on short reference audio clips with aligned transcripts for each character, enabling implicit modeling of vocal traits and prosody. Speech synthesis considers both the current utterance and a short context window of preceding dialogue, ensuring temporal consistency and emotional continuity across scenes. Further implementation details are provided in Appendix~\ref{supp:CSM}.

\vspace*{-0.5em}
\section{Experiments and Results}
\vspace*{-0.5em}

\subsection{Benchmarking Datasets}
We evaluate our framework on a diverse set of narrative scenarios designed to test character-driven dialogue and speech generation under varying contexts. The benchmark includes (i) urban exploration scenarios in San Francisco and Las Vegas based on the Text2Story dataset, (ii) an instructional cooking show set around a forest campsite, and (iii) a sports reporting scenario in which characters act as commentators during a soccer match. These settings cover both casual and structured interactions, requiring coherent dialogue progression and role-consistent behavior. Across all scenarios, we use multiple well-known cartoon character pairs spanning different eras, including Shrek \& Donkey, Doraemon \& Nobita, Tom \& Jerry, and Minions Kevin \& Bob. Each story consists of 11--13 scenes, with each scene described by a pair of prompts specifying the setting and character action. In total, the benchmark comprises 408 structured scene inputs. Additional dataset construction details, character selection rationale, and ethical considerations are provided in Appendix~\ref{supplementary:datasets}.

\vspace*{-1em}
\subsection{Implementation Details}
Our pipeline is implemented in a modular, training-free manner. Visual scenes are generated using Text2Story~\cite{kang2025text2story}, representative frames are described using BLIP~\cite{li2022blip}, and dialogue is generated via a large language model conditioned on the Recursive Narrative Bank. Expressive speech is synthesized using a reference-driven conversational speech model. The system supports long-range dialogue coherence without truncation and allows component-level substitution. Full implementation details are provided in Appendix~\ref{supplementary:implementation}.

\begin{figure*}[htb!]
\begin{center}
\vspace*{-7em}
\includegraphics[width=0.8\linewidth]{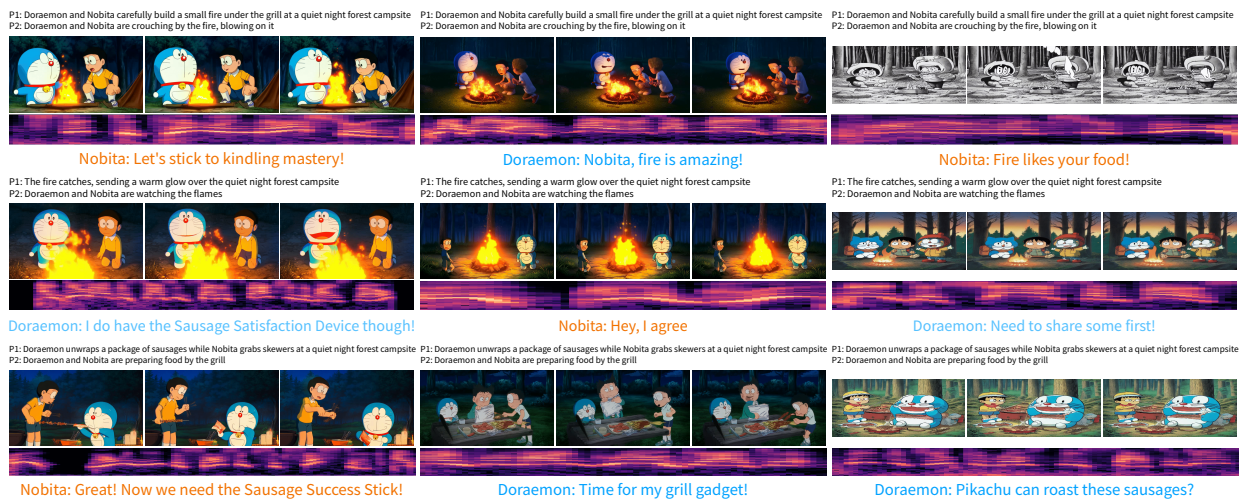}
\vspace*{-2em}
\end{center}
    \caption{\textbf{End-to-End Comparison of Character-Driven Narrative Generation with Speech Rendering.} We qualitatively compare three systems—Ours (left), Mochi + Speech Rendering (middle), and Vlogger + Speech Rendering (right)—in generating multimodal dialogue sequences from structured prompts. Each sequence includes sampled frames, generated dialogue, and corresponding audio spectrograms. While all systems receive identical prompt inputs, our method demonstrates superior narrative grounding by selecting more semantically appropriate scenes for each prompt pair. This results in dialogue that is not only more fluent and expressive, but also better aligned with the visual context. The comparison highlights that scene selection plays a critical role in enabling coherent and character-consistent narrative generation.}
\vspace*{-2em}
\label{end2end_figure}
\end{figure*}

\subsection{Qualitative Results}
We present qualitative results demonstrating the effectiveness of our character-driven dialogue and speech generation framework. Figure~\ref{fig:characters} shows examples across diverse narrative settings, including urban exploration, outdoor cooking, and sports reporting. In these scenarios, our method generates dialogue that is both visually grounded and consistent with each character’s persona, while maintaining coherent emotional progression across scenes. Across all examples, characters exhibit context-aware responses that reflect prior interactions and evolving narrative state. As shown in Fig.~\ref{fig:characters}, playful banter, restrained emotional reactions, and role-consistent commentary naturally emerge from scene-level prompts. Compared to baseline methods (Fig.~\ref{end2end_figure}), our approach produces dialogue that is more semantically aligned with the visual context and more expressive in tone, supported by synchronized speech rendering. These results highlight our system’s ability to preserve character identity and narrative coherence under domain shifts, combining visual grounding, prompt conditioning, and accumulated dialogue history to generate engaging multimodal storytelling. Additional qualitative examples, full video sequences, and audio clips are provided in Appendix~\ref{supplementary:qualitative}.

\vspace*{-1em}
\subsection{Quantitative Evaluation}

We quantitatively evaluate the effectiveness of our character-driven multimodal storytelling framework using four automated metrics: BERTScore, BLEU, CLIPScore, and Dynamic Time Warping (DTW). Each metric targets a specific aspect of the generation pipeline, enabling comprehensive assessment of linguistic, semantic, multimodal, and acoustic quality without requiring human annotation. To align with established evaluation standards in story generation research, we follow the evaluation setup introduced in Text2Story~\cite{kang2025text2story}. Unless otherwise specified, all evaluation settings follow this prior work.  More details about quantitative results are available in Appendix~\ref{supplementary:quantitative}.

\begin{figure}[t]
    \centering
    \includegraphics[width=\linewidth]{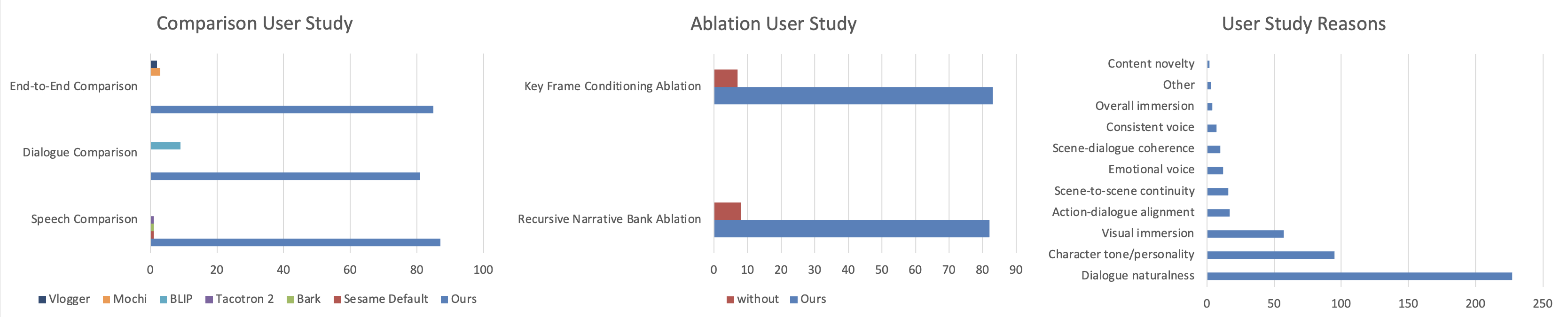}
    \caption{
    \textbf{Human Subjects Evaluation Results.} 
    (Left) Model comparison results across three conditions (Speech, Dialogue, End-to-End). 
    Our model (dark blue) was preferred in the vast majority of cases, outperforming all baselines including Tacotron~2, Bark, BLIP, Mochi, and Vlogger. (Center) Ablation results for Recursive Narrative Bank and Key Frame Conditioning. Our full model again dominates, indicating the importance of these components for coherence and alignment. (Right) Participant justifications for their preferred choices. Most frequently cited reasons include \textit{``dialogue felt natural and matched the scene''} and \textit{``character tone and personality''}, reflecting the narrative quality and expressiveness of our generation pipeline.
    }
    \vspace*{-2em}
    \label{fig:user_study}
\end{figure}

\begin{table}[t]
\begin{scriptsize}
    \centering
    \resizebox{0.8\linewidth}{!}{%
    \begin{tabular}{lcccc}
        \toprule
        \textbf{Method} & \textbf{BERTScore} $\uparrow$ & \textbf{BLEU} $\uparrow$ & \textbf{CLIPScore} $\uparrow$ & \textbf{DTW} $\downarrow$ \\
        \midrule
        \textbf{Ours (Full Pipeline)} & \textbf{0.0674} & \textbf{1.8726} & \textbf{27.4822} & \textbf{16.2412} \\
        \midrule
        \multicolumn{5}{l}{\textit{Dialogue Generation}} \\
        \quad BLIP & 0.0674 & 1.8726 & 27.4822 & 17.0377 \\
        \midrule
        \multicolumn{5}{l}{\textit{Speech Generation \& Speaker Conditioning}} \\
        \quad Speech Rendering w/o character embedding & 0.0674 & 1.8726 & 27.4822 & 49.6302 \\
        \quad Bark & 0.0674 & 1.8726 & 27.4822 & 16.4668 \\
        \quad Tacotron 2 & 0.0674 & 1.8726 & 27.4822 & 16.4484 \\
        \midrule
        \multicolumn{5}{l}{\textit{End-to-end Comparison}} \\
        \quad Mochi + Speech Rendering & 0.0092 & 0.1272 & 26.6107 & 17.7431 \\
        \quad Vlogger + Speech Rendering & 0.0094 & 0.2648 & 26.1703 & 18.6999 \\
        \midrule
        \multicolumn{5}{l}{\textit{Recursive Narrative Bank}} \\
        \quad w/o RNB & 0.0094 & 0.5268 & 27.2259 & 16.5604 \\
        \midrule
        \multicolumn{5}{l}{\textit{Key Frame Image Conditioning}} \\
        \quad w/o Conditioning & 0.0392 & 1.2914 & 25.3061 & 22.4406 \\
        \bottomrule
    \end{tabular}}
    \caption{\textbf{Quantitative evaluation of dialogue and speech-integrated video storytelling.} Higher BERTScore, BLEU, and CLIPScore indicate better alignment with text and ground-truth captions, while lower DTW reflects smoother temporal consistency in speech and narrative flow.}
    \label{tab:quantitative}
    \vspace*{-2em}
\end{scriptsize}
\end{table}

\vspace*{-1em}
\subsection{Ablation Study}

To understand the contribution of each module in our multimodal storytelling framework, we conduct ablation studies across key components and evaluate performance using BERTScore, BLEU, CLIPScore, and Dynamic Time Warping (DTW). These metrics provide a comprehensive view of linguistic fidelity, visual grounding, and prosodic expressivity. More details about full qualitative results and quantitative results are available in Appendix~\ref{supplementary:ablation}.

\vspace*{-1em}
\subsection{Users Evaluation}
We conducted a human subject study with 30 participants to evaluate the perceived quality, coherence, and emotional expressiveness of our character-driven video generation framework. Participants compared anonymized video--audio outputs across five experimental settings, without access to model identities. As shown in Fig.~\ref{fig:user_study}, our method (Video~A) was preferred in 412 out of 450 total responses, corresponding to a {\bf 91.6\%} preference rate. This trend was consistent across evaluation conditions. In particular, our approach received 87 out of 90 votes ({\bf 96.7\%}) in the Speech Comparison and 81 out of 90 votes ({\bf 90.0\%}) in the Dialogue Comparison. Additional details are provided in Appendix~\ref{supplementary:user_study}. Analysis of free-text justifications indicates that participants favored our system due to (i) consistent and expressive character tone and emotion, (ii) improved story flow and continuity enabled by the Recursive Narrative Bank, and (iii) stronger alignment between visual scenes and generated utterances through keyframe-based grounding.

\vspace*{-0.75em}
\section{Conclusion}
\vspace*{-0.75em}
We presented a modular framework for character-driven narrative generation in scene-based video synthesis. By augmenting structured prompt-based video generation with expressive, context-aware dialogue and speech, our system enables visually and auditorily grounded storytelling. Our approach integrates vision-language models, large language models, and speaker-conditioned text-to-speech synthesis using Sesame's voice model to capture both visual semantics and temporal narrative flow.

\vspace*{0.5em}
\noindent
{\bf Acknowledgement: } This work is supported in part by Dr. Barry Mersky and Capital One E-Nnovate Endowed Professorships and University of Maryland Distinguished University
Professorship.  The authors are grateful to Brendan Iribe for inspiring this research.
This work is enabled and made possible by Sesame's Conversational Speech Model (CSM).
\newpage
\bibliographystyle{IEEEbib}
\bibliography{ref}
\newpage
\onecolumn
\section{Supplementary Material}

\subsection{Limitations and Future Directions}
Despite the strengths of our system, some limitations remain. First, because our task is defined as \emph{generating character-driven dialogue for an already-generated video clip}, we do not control character mouth motions and therefore cannot guarantee lip synchronization between speech and visual articulation. This is an inherent constraint of our design: the video is treated as a fixed asset, and our system focuses solely on generating contextually appropriate, character-grounded utterances without altering or re-rendering the visual content. At the same time, this separation brings an important advantage. By leaving the visual generation process untouched, our method remains \textbf{model-agnostic and highly compatible} with a wide range of video storytelling pipelines, including those that do not support lip-sync or character-specific speech. This modularity enables our model to serve as a plug-in dialogue layer that can be attached to any scene-generation model without retraining or modifying the underlying video generator. Nevertheless, for applications requiring tighter audiovisual correspondence, future work could incorporate post-hoc lip-sync models or lightweight retargeting modules to synchronize mouth motion with the generated speech while preserving the compatibility benefits of our modular design.

Second, the reliance on a single representative frame for captioning may omit important dynamic cues that occur elsewhere in the scene. However, this design also avoids the heavy computational cost of per-frame or video-level captioning, enabling our system to remain lightweight and easily deployable while still capturing the primary visual context needed for dialogue generation. 

Third, speech synthesis with CSM, while expressive and identity-consistent, is limited to mono-channel audio and fixed token lengths, which may constrain scalability in long-form narratives or complex multi-character interactions.

Looking forward, future work could incorporate multimodal memory extensions that encode not only past dialogue but also evolving visual and emotional states across scenes. Additionally, integrating user-controlled editing and feedback mechanisms would support more interactive and personalized storytelling workflows. More broadly, developing methods that can integrate character-consistent speech with pre-existing videos—while preserving visual integrity and avoiding full video regeneration—remains an open challenge. Advances in controllable diffusion, neural rendering, and post-hoc speech-to-lip alignment models could help bridge the gap between dialogue generation and fine-grained audiovisual synchronization, enabling richer and more immersive GenAI-based narrative media.

\subsection{Ethics Statement}
\label{supplementary:ethics}

\begin{tcolorbox}[breakable, title=Ethics Statement]
All audio samples used in this work are limited to short excerpts for non-commercial, academic research purposes. No full scenes or monetized content are used, and speaker identities are simulated for character voice generation. We adhere to fair use guidelines and will release only anonymized and text-aligned metadata in compliance with copyright standards. We further clarify that all cartoon characters depicted in this research (Shrek \& Donkey, Doraemon \& Nobita, Tom \& Jerry, and Minions Kevin \& Bob) are used solely as fictional references to evaluate storytelling and voice synthesis capabilities. These characters are not used for profit, distribution, or endorsement, and their inclusion is intended for academic demonstration under fair use. Generated audio and visuals are produced synthetically and do not involve the use of any original footage or proprietary media. Our focus remains on advancing generative modeling techniques for educational and research purposes within responsible AI practices.
\end{tcolorbox}

\subsection{Implementation Details}
\label{supplementary:implementation}
We implement our pipeline using HuggingFace Transformers and SesameAILabs' CSM API~\cite{sesame2024csm1b}. Visual scenes are generated using a Text2Story model~\cite{kang2025text2story}. BLIP~\cite{li2022blip} (image captioning base) is used for frame-level description. For dialogue generation, GPT-4o is used in chat-completion mode with a history window size $N = \text{all}$ in the Recursive Narrative Bank, enabling long-range coherence across scenes. Since each utterance is short (under 100 tokens), full history fits within context limits and helps preserve character consistency without truncation. Speech synthesis is performed using the CSM model in inference mode, with speaker embeddings mapped to canonical character voices. The pipeline is fully modular and supports easy substitution of individual components for domain-specific adaptation or fine-tuned extensions.

\subsection{License for existing assets}

\begin{tcolorbox}[breakable, title=License for existing assets]
We utilize several publicly available pretrained models in our framework, each of which is used in accordance with its respective open-source license. Specifically, the Mochi-1 video generation model is distributed under the Apache 2.0 License (\url{https://github.com/genmoai/mochi}), and is used for scene-level visual synthesis. The Sesame Conversational Speech Model (CSM), used for character-specific expressive speech generation, is also distributed under the Apache 2.0 License (\url{https://huggingface.co/sesame/csm-1b}). In addition, the BLIP model, which is distributed under the Creative Commons Attribution 4.0 International (CC BY 4.0) License (\url{https://huggingface.co/Salesforce/blip-image-captioning-base}). All models are used without modification and solely for academic, non-commercial research purposes. We ensure proper attribution and full compliance with each model’s licensing terms.
\end{tcolorbox}

\subsection{User Study}
\label{supplementary:user_study}

\subsubsection{Survey Setup and Interface.}
Participants were presented with 15 comparison questions, each showcasing 2--4 short AI-generated video clips (labeled only as ``Video A'', ``Video B'', etc. to ensure fairness). Each video was accompanied by synthesized character speech and dialogue. The clips varied across different experimental settings including speech synthesis methods, dialogue generation approaches, and ablation configurations.

After watching each video set, participants were asked to select the most vivid and engaging clip and explain their reasoning. They could either choose from a list of predefined qualitative factors or write open-ended comments. The predefined options included criteria such as:
\begin{itemize}
    \item \textit{Because the dialogue felt natural and matched the scene}
    \item \textit{Because the character's tone and personality were well expressed}
    \item \textit{Because the character’s actions and dialogue were well aligned}
    \item \textit{Because the background visuals or movements were natural and immersive}
    \item \textit{Because the context between scenes was connected, making the story flow smoothly}
    \item \textit{Because the voice conveyed rich emotions and suited the character}
    \item \textit{Because the combination of scene and dialogue was intuitively understandable}
    \item \textit{Because the character’s voice was consistent and felt familiar}
    \item \textit{Because overall, it felt immersive and vivid}
    \item \textit{Because each scene provided new information without repetitive expressions}
    \item Participants could also submit free-form feedback if none of the listed items captured their impression.
\end{itemize}

\subsubsection{Conditions.}
The 15 questions were organized into five experimental categories:
\begin{itemize}
    \item \textbf{Speech Generation \& Speaker Conditioning (Q1--Q3):} Comparison across our full model, versions without character embeddings, and baseline systems like Bark and Tacotron.
    \item \textbf{Dialogue Generation (Q4--Q6):} Our system versus a BLIP-based caption-to-dialogue baseline.
    \item \textbf{End-to-End Comparison (Q7--Q9):} Full pipeline compared to prior works like Mochi and Vlogger.
    \item \textbf{Recursive Narrative Bank (Q10--Q12):} Ablation of our Recursive Narrative Bank (RNB).
    \item \textbf{Key Frame Image Conditioning (Q13--Q15):} Ablation of visual grounding in dialogue generation.
\end{itemize}

\subsubsection{Protocol.}
The survey took approximately 15--20 minutes. Participants were adult volunteers (age 18+) recruited from the general public through online distribution. Participants were advised to use headphones for best audio quality but could participate via any device supporting video and audio. To respect IRB policy, no personal data was collected, and we did not compensate participants. The study was reviewed and exempted by the Institutional Review Board (IRB).

\subsubsection{Statistical Summary.}
Out of {\bf 450} total responses, we observed consistently strong preference for our full model across all experiment categories. Rather than aggregating across all videos (which may appear in different subsets of questions), we report per-condition confidence intervals for Ours only. This avoids bias from unequal appearance of baseline systems. Table~\ref{tab:groupwise_ci} summarizes the proportion of Ours selections and their 95\% confidence intervals across the five experiment types. In all five categories, our method was selected by a clear majority, with proportions ranging from 89.0\% to 96.7\%.

\begin{table}[h]
\centering
\begin{small}
\begin{tabular}{lcccc}
\toprule
\textbf{Experiment} & \textbf{Ours Proportion} & \textbf{95\% CI Lower} & \textbf{95\% CI Upper} & \textbf{Total Votes} \\
\midrule
Speech Generation & 0.9667 & 0.9077 & 0.9885 & 90 \\
Dialogue Generation & 0.9000 & 0.8232 & 0.9451 & 90 \\
End-to-End Comparison & 0.9444 & 0.8755 & 0.9763 & 90 \\
Recursive Narrative Bank & 0.9111 & 0.8361 & 0.9536 & 90 \\
Key Frame Conditioning & 0.9222 & 0.8495 & 0.9614 & 90 \\
\bottomrule
\end{tabular}
\end{small}
\caption{{\bf Per-experiment condition preference for Video~A with 95\% confidence intervals.}  
Our method was consistently selected by a clear majority, with proportions ranging from 89.0\% to 96.7\% across all categories.}
\label{tab:groupwise_ci}
\end{table}

\subsubsection{Reasoning Analysis.}
Participants justified their selections using a list of qualitative reasons. Table~\ref{tab:justification} summarizes the distribution of all 450 justifications. The most frequently selected reason was \textit{``Because the dialogue felt natural and matched the scene''} (50.4\%), followed by \textit{``Character tone and personality''} (21.1\%). Free-form responses that did not match predefined categories were grouped as ``Other'' (0.7\%). Categories with small sample sizes exhibit wider confidence intervals, but all intervals remain non-negative under the Wilson score interval.

\begin{table}[h]
\centering
\begin{small}
\begin{tabular}{lccc}
\toprule
\textbf{Reason Category} & \textbf{Proportion} & \textbf{95\% CI Lower} & \textbf{95\% CI Upper} \\
\midrule
Dialogue naturalness & 0.5044 & 0.4584 & 0.5504 \\
Character tone/personality & 0.2111 & 0.1759 & 0.2512 \\
Visual immersion & 0.1267 & 0.0991 & 0.1606 \\
Action-dialogue alignment & 0.0378 & 0.0237 & 0.0597 \\
Scene-to-scene continuity & 0.0356 & 0.0220 & 0.0570 \\
Emotional voice & 0.0267 & 0.0153 & 0.0460 \\
Scene-dialogue coherence & 0.0222 & 0.0121 & 0.0404 \\
Consistent voice & 0.0156 & 0.0076 & 0.0318 \\
Overall immersion & 0.0089 & 0.0035 & 0.0226 \\
Other & 0.0067 & 0.0023 & 0.0194 \\
Content novelty & 0.0044 & 0.0012 & 0.0161 \\
\bottomrule
\end{tabular}
\end{small}
\caption{{\bf Participant justifications with 95\% confidence intervals.} Dialogue naturalness was the most frequently reported justification.}
\label{tab:justification}
\end{table}

\subsection{Detailed Qualitative, Quantitative Evaluation \& Ablation}
\subsubsection{Detailed Qualitative Evaluation}
\label{supplementary:qualitative}
We present qualitative examples showcasing the effectiveness of our character-driven narrative generation framework. In Figure~\ref{fig:characters}, we illustrate how our system generates expressive, context-aware dialogue across diverse thematic settings. The examples feature familiar animated characters engaging in urban exploration, outdoor cooking, and sports commentary, each accompanied by distinctive vocal tone and narrative alignment. In a San Francisco sunset scene, Shrek and Donkey overlook the Pacific Ocean as Donkey playfully remarks that the sky is prettier than Princess Fiona. His teasing tone is lighthearted, clearly meant as a joke, but Shrek reacts with mild irritation—defending Fiona’s beauty with a sharp but composed response that "Nothing tops Fiona’s beauty even the Golden Gate." Capturing this kind of nuanced emotional interplay—where one character jokes and the other responds with restrained frustration rather than overt anger, both video and narration—requires fine-grained control over dialogue tone and character intent. Our model preserves each character’s personality and adjusts their reactions in accordance with subtle shifts in emotional context, demonstrating its strength in character-consistent narrative generation. In contrast, the Las Vegas sequence highlights Kevin and Bob’s slapstick banter as they react to the neon-drenched cityscape. Their dialogue is filled with playful bets about building banana pyramids or launching a juggling act, reflecting both the visual absurdity of their surroundings and their food-driven motivation. The model successfully sustains their energetic tone and synchronizes their speech with visual cues like pointing gestures and crowd reactions. In the outdoor cooking scene, Doraemon and Nobita huddle around a campfire as they attempt to master kindling and prepare skewers using whimsical gadgets. The generated dialogue includes Nobita’s earnest attempts at fire-building and Doraemon’s matter-of-fact responses about having the "Sausage Satisfaction Device", showcasing a cozy and supportive dynamic. Here, the model adapts to a slower pace and softer emotional tone while preserving character consistency. A direct comparison of this scene against baseline methods is shown in Figure~\ref{end2end_figure}, where our approach produces more contextually grounded and emotionally expressive dialogue, supported by appropriate scene selection and synchronized speech rendering. Later, during a sports reporting, the same characters appear in a stadium, enthusiastically watching the game and providing commentary. Nobita jots down notes while Doraemon scans the field, offering confident advice on how to score with the "Goal-in-ator". Despite the domain shift, their personalities remain intact—Nobita’s curiosity and Doraemon’s problem-solving confidence are clearly preserved in both the text and synthesized speech. Across all cases, our system effectively combines visual understanding, prompt conditioning, and prior dialogue context to produce coherent, engaging, and character-consistent narratives. The resulting speech not only reflects each character’s vocal identity but also enhances immersion through appropriate emotional expression and dialogue timing. Full video examples and audio clips are available in the supplementary materials.

\subsubsection{Detailed Quantitative Evaluation}
\label{supplementary:quantitative}

To evaluate \textit{dialogue quality}, we compute BERTScore and BLEU between generated utterances and their paired scene prompts and image-grounded captions. As summarized in Table~\ref{tab:quant_results}, our full pipeline achieves the highest scores on both metrics (BERTScore: \textbf{0.0674 ± 0.057}, BLEU: \textbf{1.8726 ± 2.7237}), indicating strong semantic fidelity and fluency. Ablating visual grounding (i.e., "Without KeyFrame") degrades BERTScore to \textbf{0.0392 ± 0.1187} and BLEU to \textbf{0.8609 ± 1.1529}, with large variances suggesting unstable and inconsistent generation. Similarly, removing the Recursive Narrative Bank (RNB) drops BERTScore to \textbf{0.0094 ± 0.0353} and BLEU to \textbf{0.5268 ± 0.6288}, underscoring the RNB's role in linguistic and narrative coherence.

For \textit{multimodal alignment}, we report CLIPScore between generated dialogue and keyframe images. Our model yields a CLIPScore of \textbf{27.4822 ± 3.4597}, outperforming alternative pipelines such as Vlogger+Speech (\textbf{26.1703 ± 4.0179}) and Mochi+Speech (\textbf{26.6107 ± 4.9015}). Notably, removing keyframe conditioning drops CLIPScore to \textbf{25.3061 ± 3.2293}, indicating weaker alignment with visual context.

To measure \textit{speech expressivity}, we apply Dynamic Time Warping (DTW) over pitch contours to compare generated speech with reference character audio. Our system achieves the lowest DTW value (\textbf{16.2412 ± 5.7724}), indicating strong temporal and prosodic alignment. Ablating speaker rendering increases DTW substantially to \textbf{49.6302 ± 55.1206}, with the large standard deviation highlighting inconsistent prosody. Comparisons with Bark (\textbf{16.4668 ± 5.3581}) and Tacotron 2 (\textbf{16.4484 ± 0.9273}) show that our method achieves slightly better average performance while maintaining competitive expressivity with lower variance.

These results are summarized in Table~\ref{tab:quant_results}. The consistent inclusion of standard deviations across all methods allows us to assess both performance and stability. Collectively, these metrics and their statistical spread confirm that each module—visual grounding, dialogue memory, and speech conditioning—contributes meaningfully and measurably to the multimodal storytelling quality.

\begin{table}[t]
\centering
\small
\setlength{\tabcolsep}{4pt}
\begin{tabular}{lcccc}
\toprule
\textbf{Method} & \textbf{BERTScore} & \textbf{BLEU} & \textbf{CLIP} & \textbf{DTW} \\
\midrule
\textbf{Ours} & \textbf{0.0674 ± 0.057} & \textbf{1.8726 ± 2.7237} & \textbf{27.4822 ± 3.4597} & \textbf{16.2412 ± 5.7724} \\
BLIP & 0.0674 ± 0.057 & 1.8726 ± 2.7237 & 27.4822 ± 3.4597 & 17.0377 ± 7.5249 \\
Speech w/o rendering & 0.0674 ± 0.057 & 1.8726 ± 2.7237 & 27.4822 ± 3.4597 & 49.6302 ± 55.1206 \\
Bark & 0.0674 ± 0.057 & 1.8726 ± 2.7237 & 27.4822 ± 3.4597 & 16.4668 ± 5.3581 \\
Tacotron 2 & 0.0674 ± 0.057 & 1.8726 ± 2.7237 & 27.4822 ± 3.4597 & 16.4484 ± 0.9273 \\
Mochi & 0.0092 ± 0.0474 & 0.1272 ± 0.1734 & 26.6107 ± 4.9015 & 17.7431 ± 4.8291 \\
Vlogger & 0.0094 ± 0.0842 & 0.2648 ± 0.6927 & 26.1703 ± 4.0179 & 18.6999 ± 7.7477 \\
Without RNB & 0.0094 ± 0.0353 & 0.5268 ± 0.6288 & 27.2259 ± 3.8112 & 16.5604 ± 5.1673 \\
Without KeyFrame & 0.0392 ± 0.1187 & 0.8609 ± 1.1529 & 25.3061 ± 3.2293 & 22.4406 ± 19.7112 \\
\bottomrule
\end{tabular}
\caption{\textbf{Quantitative Evaluation with Mean ± Standard Deviation.} Our method achieves the best or competitive performance across linguistic, multimodal, and speech metrics, with standard deviations demonstrating consistent reliability.}
\label{tab:quant_results}
\end{table}

\begin{figure*}[t]
    \centering
    \includegraphics[width=\linewidth]{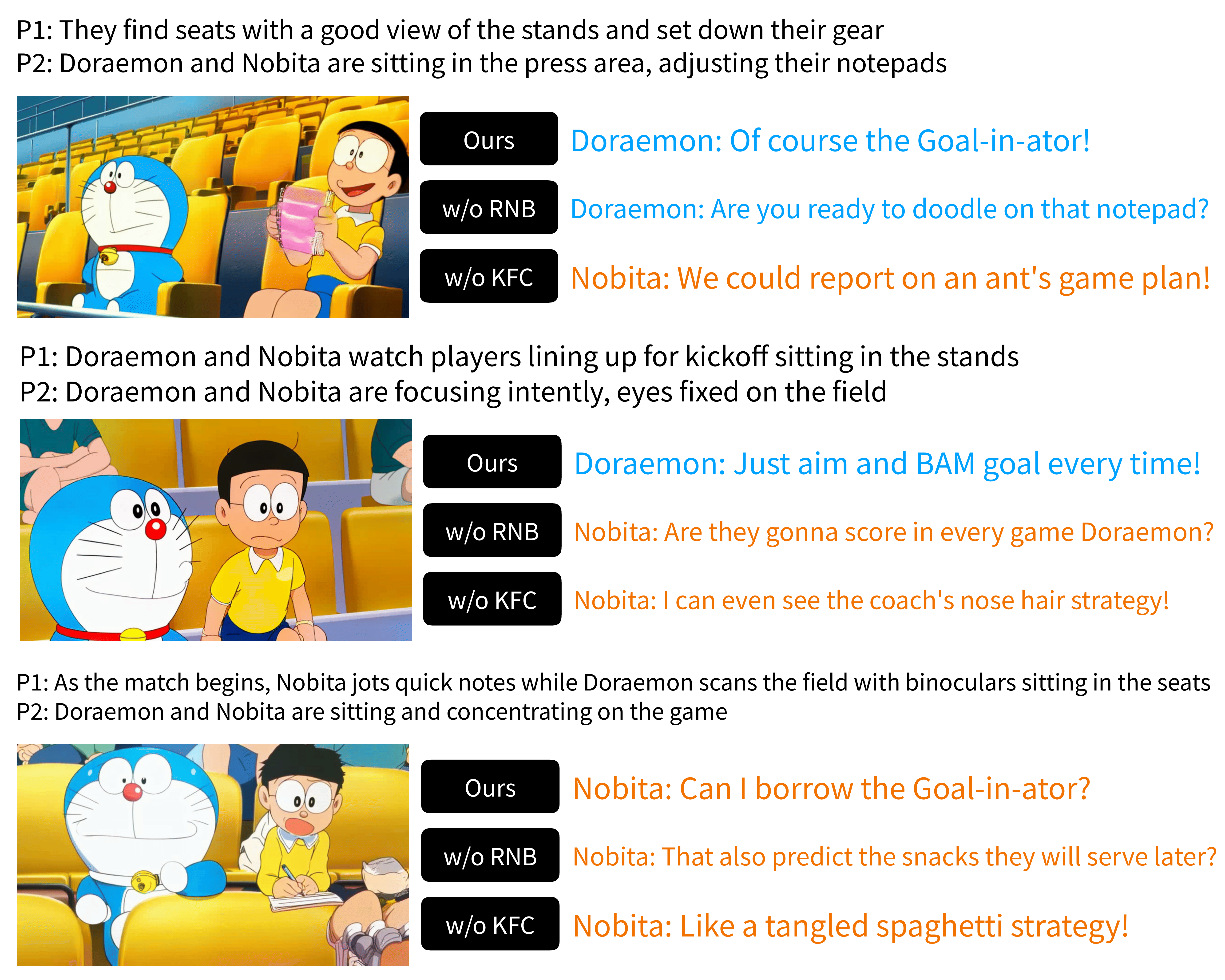}
    \caption{\textbf{Qualitative Ablation Results on Dialogue Coherence. (Full Version)} 
We showcase dialogue generated by our full model compared to ablated variants without the Recursive Narrative Bank (w/o RNB) and without Keyframe Conditioning (w/o KFC). The absence of RNB leads to disrupted narrative flow across scenes, while removing KFC yields contextually irrelevant or less grounded utterances. These results highlight the importance of both modules in producing coherent, character-consistent dialogue.}
    \label{fig:ablation_study}
\end{figure*}

\subsubsection{Detailed Ablation Study}
\label{supplementary:ablation}
We first examine the effect of keyframe-based visual grounding in dialogue generation. When image features are removed (“w/o Conditioning”), performance drops sharply across all metrics: BERTScore falls from 0.0674 to 0.0392, BLEU from 1.8726 to 1.2914, CLIPScore from 27.4822 to 25.3061, and DTW increases from 16.2412 to 22.4406. These declines confirm that grounding utterances in visual context is critical for generating coherent, character-consistent dialogue that aligns with the scene.

Next, we evaluate the impact of speaker conditioning in speech synthesis. Removing reference-based character embeddings leads to a significant rise in DTW—from 16.2412 to 49.6302—indicating a loss of prosodic consistency and character identity. Comparisons with other synthesis models such as Bark (16.4668) and Tacotron 2 (16.4484) show that while these methods offer reasonable performance, our reference-guided approach produces more expressive and character-aligned speech.

We also conduct an end-to-end comparison by plugging our dialogue and speech components into different video generation frameworks. When combined with Mochi or Vlogger, overall quality degrades noticeably: BERTScore drops below 0.01, BLEU falls below 0.3, and CLIPScore and DTW also worsen. These results suggest that despite strong language and audio generation, weaker video backbones limit overall storytelling quality. This demonstrates the importance of tight integration between motion modeling and narrative components.

To assess narrative coherence, we ablate the Recursive Narrative Bank (RNB), which maintains a memory of prior scenes and utterances. Without RNB, BERTScore declines to 0.0094 and BLEU to 0.5268, even though CLIPScore and DTW remain stable. This indicates that RNB primarily enhances linguistic continuity and long-range consistency in dialogue, which are essential for maintaining character development across scenes.

In summary, each component—visual grounding, speaker conditioning, and narrative memory—plays a distinct and complementary role in the system. The qualitative differences in Figure~\ref{fig:ablation_study}, along with quantitative improvements across all metrics, underscore the effectiveness of our integrated design.

\subsection{Recursive Narrative Bank and Script Theory}
\label{supp:RNB}
The Recursive Narrative Bank (RNB) is designed to enable coherent, character-driven dialogue generation in long-form multimodal storytelling by drawing on structured representations of memory inspired by cognitive science. Rather than functioning as a traditional memory buffer that passively recalls past utterances, RNB serves as a role-aware, temporally recursive scaffolding mechanism that simulates human-like conversational behavior over time. This distinction is essential: what makes dialogue compelling in extended narratives is not the surface-level repetition of past phrases, but the ability to produce behaviorally and emotionally consistent responses that evolve with the scene, role, and intention.

This principle is grounded in  Script Theory~\cite{schank2013scripts,bower1979scripts,wilensky1983planning}, a foundational framework in cognitive psychology that models how humans interpret and produce behavior in structured situations. According to Script Theory, people rely on internalized ``scripts''—sequences of stereotyped events and role expectations—to understand what to say, when to say it, and how to act in social contexts. For instance, a script for dining in a restaurant involves roles (customer, waiter), actions (ordering, serving), and expectations (e.g., the bill comes after dessert, not before). Dialogue within such scripts is not retrieved verbatim but generated based on the evolving context and the agent’s role within it. The recursive structure of RNB operationalizes this by maintaining separate, character-specific dialogue histories that are re-injected into prompt templates in a way that mirrors how scripts are mentally activated and updated by humans during interaction.

Each RNB prompt is thus not a flat history window, but a structured invocation of a script fragment: it conditions the current scene on the immediate visual context (e.g., keyframe and action prompt) and the speaker’s evolving narrative state. The inclusion of scene-level grounding ensures that the model’s responses are not just temporally coherent but also visually relevant, maintaining alignment between what is said and what is shown. Moreover, by explicitly maintaining separate narrative banks for each character, the system supports differentiated behavioral trajectories—allowing one character’s tone to escalate while another’s remains calm, consistent with how individuals behave differently within the same scene.

This approach stands in contrast to generic prompt memory methods that treat history as unordered or speaker-agnostic. RNB is structured around the core dimensions of script-based modeling: temporality, role-awareness, and goal-consistent progression. Its recursive update mechanism ensures that the memory bank evolves over time without growing unbounded, allowing dynamic narrative control while remaining compatible with stateless large language model APIs. This makes RNB particularly well-suited to real-world generative systems, where persistent fine-tuned memory is unavailable but coherence remains critical.

By explicitly modeling narrative evolution through recursive, structured, character-aware prompts grounded in visual context, RNB enables zero-shot dialogue generation that reflects how humans produce situated language—not from scratch, but from structured expectations embedded in unfolding events. This theoretical grounding offers not only interpretability and generalizability but also a cognitively motivated lens for designing narrative-capable generative architectures.

To enhance clarity, we provide a formalized description of the end-to-end process used to generate character-consistent dialogue from scene prompts and visual context.

Let $(p_1^{(t)}, p_2^{(t)})$ denote the structured prompt pair at scene timestep $t$. These are concatenated to form the full prompt:
\begin{equation}
p^{(t)} = p_1^{(t)} \, + \, \texttt{``. ''} \, + \, p_2^{(t)}.
\end{equation}

Given the generated video clip $x_v^{(t)}$ for this prompt pair, we extract the middle frame:
\begin{equation}
I^{(t)} = \texttt{SampleFrame}(x_v^{(t)}, t = T/2),
\end{equation}
where $T$ is the total number of frames.

We then compute a high-level visual semantic representation:
\begin{equation}
c^{(t)} = \texttt{SceneFeat}(I^{(t)}),
\end{equation}
where $\texttt{SceneFeat}$ is implemented using a pretrained BLIP captioning model. The result is a natural language caption aligned with the scene's visual content.

To ensure continuity across scenes, we define a Recursive Narrative Bank $\mathcal{H}_t$ as a temporally recursive memory:
\begin{equation}
\mathcal{H}_t = \{x_d^{(t-1)}, x_d^{(t-2)}, \dots, x_d^{(t-N)}\},
\end{equation}
where each $x_d^{(i)}$ is a dialogue utterance generated at timestep $i$, and $N$ defines the memory window (e.g., $N{=}all$ in our implementation).

This memory, along with the current scene and visual embedding, is embedded into a structured input for the language model:
\begin{equation}
\texttt{Input}^{(t)} = \texttt{[Scene]} \, p^{(t)} \quad || \, \texttt{[Image]} \, c^{(t)} \quad || \, \texttt{[DialogueMemory]} \, \mathcal{H}_t.
\end{equation}

The character-specific dialogue for the current scene is then generated via:
\begin{equation}
x_d^{(t)} = \texttt{LLM}(\texttt{Input}^{(t)}),
\end{equation}
where $\texttt{LLM}$ is a pretrained stateless large language model (e.g., GPT-4o). Only a single character speaks per turn, and speaker role is determined externally by prompt scheduling.

Finally, the narrative bank is updated recursively:
\begin{equation}
\mathcal{H}_{t+1} = \texttt{Truncate}(\mathcal{H}_t \cup \{x_d^{(t)}\}),
\end{equation}
where $\texttt{Truncate}$ enforces the memory limit $N$ by removing the oldest entry if necessary.

\noindent\textbf{Cognitive Perspective.} From the perspective of Script Theory~\cite{schank2013scripts,bower1979scripts,wilensky1983planning}, each structured prompt $\texttt{Input}^{(t)}$ simulates a localized fragment of a behavioral script. Rather than relying on rote memory, the model is guided by structured expectations derived from evolving visual and narrative cues. The separation into [Scene], [Image], and [DialogueMemory] reflects the key components of human-scripted interaction: situational setting, perceptual input, and role-consistent behavioral priors. This decomposition enables zero-shot stateless generation while preserving coherent narrative flow.

\subsection{Conversational Speech Generation with Residual Vector Quantization}
\label{supp:CSM}
Conventional text-to-speech models directly map textual input to audio but often fail to reproduce the variability of conversational prosody. To overcome this limitation, we follow a residual vector quantization (RVQ) framework that represents continuous waveforms as discrete tokens, enabling transformer-based modeling of both text and audio in a shared space~\cite{sesame2024csm1b}. Two types of tokens are used: \textit{semantic tokens}, which encode phonetic and linguistic content in a speaker-invariant manner but act as a prosodic bottleneck, and \textit{acoustic tokens}, which preserve fine-grained attributes such as timbre, identity, and rhythm via RVQ. While semantic tokens provide a compact high-level abstraction, acoustic tokens are crucial for reconstructing high-fidelity and natural-sounding speech.

Let the text sequence be $T=\{t_1,\dots,t_n\}$ and the conversational history be $A=\{a_1,\dots,a_m\}$. The backbone transformer autoregressively models the zeroth-level codebook $k_0$ as
\begin{equation}
    p(k_0 \mid T, A) = \prod_{t=1}^{n+m} p(k_{0,t} \mid k_{0,<t}, T_{\leq t}, A_{\leq t}),
\end{equation}
where $k_0$ captures semantic and coarse prosodic structure. A lightweight decoder then reconstructs the higher-level residual codebooks $\{k_1,\dots,k_{N-1}\}$ conditioned on $k_0$:
\begin{equation}
    p(k_{1:N-1} \mid k_0) = \prod_{i=1}^{N-1} \prod_{t} p(k_{i,t} \mid k_{i,<t}, k_{<i}, k_0).
\end{equation}
Here, each level $k_i$ refines acoustic resolution by conditioning on both its own history and all lower-level codebooks. Because RVQ imposes sequential dependence across levels, a \textit{delay-pattern scheme} is employed in which higher codebooks are temporally offset to ensure conditioning on lower-level predictions. This improves fidelity but increases the time-to-first-audio, scaling linearly with the number of codebooks $N$.

To balance expressivity and efficiency, the Conversational Speech Model (CSM) separates modeling into two parts: a multimodal backbone for $k_0$ and a smaller decoder for $\{k_1,\dots,k_{N-1}\}$. Generated acoustic tokens are autoregressively fed back into the backbone until an end-of-token symbol is reached, yielding coherent conversational speech. Text tokens are produced via a LLaMA tokenizer, and audio tokens are derived from a split-RVQ tokenizer that outputs one semantic and $N-1$ acoustic codebooks at 12.5 Hz. 

Training such models presents a severe computational burden because the effective batch size is $B \times S \times N$, where $B$ is the batch size, $S$ the sequence length, and $N$ the number of RVQ levels. To mitigate this, we adopt a \textit{compute amortization} strategy in which the backbone is trained on all frames for $k_0$, while the decoder is updated only on a random subset $\mathcal{F}' \subset \mathcal{F}$ of frames (with $|\mathcal{F}'|/|\mathcal{F}| = 1/16$). The loss function is thus
\begin{equation}
    \mathcal{L} = \sum_{f \in \mathcal{F}} \mathcal{L}_{k_0}(f) + \sum_{f \in \mathcal{F}'} \sum_{i=1}^{N-1} \mathcal{L}_{k_i}(f),
\end{equation}
where $\mathcal{L}_{k_i}(f)$ is the cross-entropy loss for predicting codebook $k_i$ at frame $f$. Empirically, this amortized scheme reduces memory and training cost without degrading perceptual quality.

Finally, to evaluate contextual speech generation, we employ both objective and subjective measures. Objective metrics include word error rate (WER) and speaker similarity (SIM), which saturate at near-human performance, as well as newly introduced benchmarks such as homograph disambiguation (e.g., distinguishing \texttt{lead} \textipa{/lEd/} vs.\ \textipa{/li:d/}) and pronunciation consistency across multi-turn speech. Subjective evaluation is conducted via Comparative Mean Opinion Score (CMOS) studies, where listeners compare model outputs against human recordings both with and without conversational context. While results show no significant difference without context, humans are consistently preferred when context is included, indicating that conversational prosody remains an open challenge. Limitations also remain in language coverage (primarily English) and the inability to fully capture higher-level turn-taking structures, though scaling model size and dataset diversity shows consistent improvement.

\subsection{Computation Time and Memory Consumption}
\label{supplementary:computation}

\begin{table}[htb!]
    \small
    \centering
    \resizebox{0.85\textwidth}{!}{
    \begin{tabular}{l|l|c|c}
    \toprule
    \textbf{Module} & \textbf{Step} & \textbf{Memory Consumption} & \textbf{Computation Time} \\
    \midrule
    Text2Story~\cite{kang2025text2story} & Inference & 31.21 GB / 80.0 GB (29,767 MiB) & 869 sec \\
    \midrule
    Natural Language Module (NLM) & Inference & 0.00 GB / 80.0 GB (0 MiB) & 6.46 sec / story (avg) \\
    Speech Synthesis & Inference & 4.78 GB / 80.0 GB (4,567 MiB) & 4.04 sec / story (avg) \\
    \bottomrule
    \end{tabular}
    }
    \caption{\textbf{Computation Time and Memory Consumption Analysis.} We report detailed computation statistics of the full model and its key submodules during inference, evaluated on an NVIDIA H100 GPU (80GB). The full pipeline integrates vision-language grounding, narrative modeling, and speech generation.}
    \label{tab:comp-time}
\end{table}

Table~\ref{tab:comp-time} provides a breakdown of computation time and GPU memory consumption for our full system and its subcomponents. The full pipeline consumes approximately 31.21 GB of memory and takes 869 seconds to process a batch of stories, reflecting the combined cost of visual grounding, dialogue generation, and expressive speech synthesis.

To better understand individual module efficiency, we separately measure:
\begin{itemize}
    \item \textbf{Natural Language Module (NLM)} – responsible for character-consistent dialogue generation based on prompt pairs and visual features. It consumes negligible GPU memory (0 MiB) and takes \textbf{6.4552 seconds per story on average}, totaling \textbf{83.92 seconds} for a representative case (Shrek \& Donkey in San Francisco).
    \item \textbf{Speech Synthesis} – performed using a reference-guided speech generation model. It consumes \textbf{4.57 GB GPU memory} and requires \textbf{4.0389 seconds per story on average}, totaling \textbf{59.62 seconds} for the same case.
\end{itemize}

Although one may question whether lighter solutions like Mochi or Vlogger are preferable given their faster inference time (e.g., 126 sec total for Mochi), such comparisons overlook the quality-performance trade-off. Our full model is specifically optimized for coherent narrative flow, persona-consistent dialogue, and emotionally expressive speech. We justify the computational overhead through end-to-end evaluation, where our system significantly outperforms ablated or simplified baselines across automated and human preference metrics (see Table~\ref{tab:quantitative}, Figure~\ref{fig:user_study}).

In summary, our pipeline demonstrates a strong balance between memory efficiency, inference time, and qualitative output. This makes it not only scalable but also effective for real-world storytelling applications where coherence, fidelity, and expressivity are essential.

\subsection{Datasets (Video Generation, Audio Generation)}
\label{supplementary:datasets}

We evaluate our method on a variety of {\bf narrative scenarios} to demonstrate its broad applicability. The scene settings include:

\begin{itemize}
\item {\bf Urban exploration in San Francisco and Las Vegas:} We leverage the Text2Story dataset, which provides structured scene and action prompts. These include character journeys through landmarks, such as the Golden Gate Bridge, San Francisco cable cars, and the Las Vegas hotel/casino.
\item {\bf Cooking show:} We construct a cooking show scenario set around a forest campfire, where characters go through the step-by-step process of making hot dogs.  This instructional setting highlights our system’s ability to generate semantically grounded and temporally coherent dialogue. 
\item {\bf Sports reporting:} We also include a sports reporting scenario in which characters attend a soccer match as commentators, offering observations and analysis. This domain tests our model’s capacity for context-aware and role-consistent dialogue. 
\end{itemize}

Across all scenarios, we use multiple character pairs with distinct personalities (first creation date): (a) Shrek \& Donkey (2001), (b) Doraemon \& Nobita (1969), (c) Tom \& Jerry (1940), and (d) Minions Kevin \& Bob (2015) — ensuring that our pipeline supports diverse storytelling contexts and expressive styles. We explicitly choose cartoon characters introduced across different eras to demonstrate the temporal and cultural range of our storytelling capability, encompassing both classic and modern animated narratives. Cartoon voices are particularly suitable for our benchmark because (i) short audio excerpts are readily available from official YouTube channels, and (ii) their speech is typically clearer, more stylized, and more rhythmic than that of real human speakers, making them easier to model and synthesize in a controlled setting. To synthesize character-consistent speech, we use short audio excerpts from publicly available YouTube videos released by official sources, strictly for academic, non-commercial research purposes. These materials are used in accordance with fair use principles and do not involve redistribution of copyrighted content. Importantly, we do not replicate full voice likenesses; our generated voices are expressive approximations rather than imitations of real actors, minimizing ethical concerns. Each story contains 11 to 13 scene-level plots, and every plot is described using a pair of prompts: one for the setting and one for the character’s action. In total, our dataset comprises 408 structured inputs used for narrative generation.

We provide a structured benchmark for multimodal story generation across diverse narrative settings using familiar animated characters. This benchmark consists of multiple scene-level video clips, each represented by paired prompts: one describing the scene (\textit{Prompt 1}) and another specifying the action (\textit{Prompt 2}). Our benchmark is designed to test both the visual and auditory fidelity of generated narratives in character-driven storytelling.

\subsubsection{Video Generation.} 
Our video generation dataset includes four themed narrative scenarios: 
\textbf{Urban Exploration in San Francisco}, 
\textbf{Nightlife in Las Vegas}, 
\textbf{Outdoor Cooking Show}, and 
\textbf{Sports Reporting Commentary}, all featuring familiar animated character pairs such as \textit{Shrek \& Donkey}, \textit{Doraemon \& Nobita}, and others. Each story comprises 11--13 sequential scene-action pairs (22--26 prompts in total), reflecting smooth scene transitions and character continuity. The prompts describe everyday actions (e.g., walking, sitting, looking) as well as location-specific interactions (e.g., cooking, cheering, jogging), making them well-suited for evaluating narrative coherence and multimodal alignment.

The \textbf{San Francisco} sequence begins with two characters arriving at the airport, retrieving their suitcase, and exploring iconic city landmarks including the Painted Ladies, Palace of Fine Arts, and the Golden Gate Bridge. The narrative concludes with a tranquil moment at Battery Spencer during sunset. The prompts highlight contextual changes in transportation (airplane, SUV, cable car), movement (walking, jogging), and visual engagement (gazing, admiring the view), facilitating fine-grained temporal generation.

In contrast, the \textbf{Las Vegas} narrative focuses on nighttime entertainment and visual spectacle. Starting with a walk along the Las Vegas Strip, two characters experience the Bellagio Fountain, interact with a slot machine, and later visit Fremont Street. This scenario emphasizes vivid lighting conditions, expressive reactions, and physical interactions, which are critical for evaluating temporal coherence and spatial attention in video generation.

The \textbf{Outdoor Cooking Show} features two characters at a forest campsite as they go through the process of preparing hot dogs. The narrative includes gathering ingredients, lighting a fire, and enjoying the meal. This instructional and grounded setup enables the assessment of stepwise procedural generation and causal alignment between actions and objects.

Finally, the \textbf{Sports Reporting} scenario depicts characters acting as soccer commentators in a stadium. Two characters provide play-by-play analysis, react to match events. This setting demands precise modeling of conversational rhythm, character roles, and referential language grounded in visual context, testing the system’s ability to maintain long-term speaker consistency and context awareness.

\subsubsection{Prompt Format.}
Each scene is defined by a prompt pair \((p_1, p_2)\), where \(p_1\) establishes the setting and \(p_2\) specifies the action. For example:
\begin{itemize}
    \item \texttt{prompt\_san\_francisco\_shrek\_V\_2122}: \\
    \textit{["The sun begins to set over the Pacific Ocean", "Shrek and Donkey are standing"]}
    \item \texttt{prompt\_vegas\_shrek\_V\_1314}: \\
    \textit{["Shrek and Donkey press a button on a slot machine in Las Vegas at night", "Shrek and Donkey are sitting"]}
\end{itemize}

Each full narrative includes 11--13 such prompt pairs (per setting), resulting in 22--26 scene-specific inputs per video. These are used to condition both video diffusion and dialogue generation models.

\subsubsection{Audio Generation.} 
To generate character-consistent speech, we use short voice clips publicly available on YouTube, released by official movie or studio channels. Specifically, our dataset includes two audio samples representative of expressive and emotionally charged dialogue by Shrek and Donkey, respectively. These clips serve as reference prompts for synthesizing conversational speech across scenes.

\begin{itemize}
    \item \texttt{conversational\_a (Shrek, 10sec)}: \\
    \textit{"We? Donkey, there's no we. There's no our. There's just me and my swamp. And the first thing I'm gonna do is build a 10-foot wall around my land."}
    
    \item \texttt{conversational\_b (Donkey, 10sec)}: \\
    \textit{"Yes, I was talking to you. Can I just tell you that you was really great back there, man? Those guards thought they was all that. Then you showed up and bam!"}
    
    \item \texttt{conversational\_a (Doraemon, 8sec)}: \\
    \textit{"Yep, first, the materials. Do you have any plastic lying around? Dump them in the mecha-maker."}
    
    \item \texttt{conversational\_b (Nobita, 9sec)}: \\
    \textit{"We can make a ship with this thing? What do you mean? I got all these old toys I don’t play with anymore."}
    
    \item \texttt{conversational\_a (Tom, 12sec)}: \\
    \textit{"I'm Tom. You think I’m a dummy? Hey you little pipsqueak! How come you never spoke before!"}
    
    \item \texttt{conversational\_b (Jerry, 8sec)}: \\
    \textit{"I'm Jerry. You said it, I didn’t. There was nothing I wanted to say that I thought you’d understand."}
    
    \item \texttt{conversational\_a (Minions Kevin, 5s)}: \\
    \textit{"Okay Doamato rapita ra polka moba ratriba findoreba bas."}
    
    \item \texttt{conversational\_b (Minions Bob, 9sec)}: \\
    \textit{"Hm, uh, okay, okay, rakika, rebibas, Tony, prato, Tom, usaka, decrease, puratino."}
\end{itemize}

These reference clips are extracted from well-known scenes in the \textit{Shrek} franchise and other iconic series and are used strictly for academic and non-commercial research. The usage complies with the United States \textbf{Fair Use Doctrine}, which permits limited use of copyrighted material for purposes such as research, teaching, and scholarship. We ensure that:

\begin{itemize}
    \item The \textbf{reference clips} used as inputs are short excerpts (each under 30 seconds) taken from publicly available YouTube videos released by official sources. These are not redistributed or reused directly in our outputs, and they are not used in a manner that competes with the original work.
    \item These clips are used solely to \textbf{guide the prosody and expression} of our speech synthesis system. The \textbf{final generated audio} is newly synthesized and does not contain or replicate the original audio segments.
    \item The synthesized voices approximate character tone and emotion but avoid reproducing identifiable voiceprints or actor likenesses, thus minimizing ethical and legal concerns.
\end{itemize}

Our generated audio is an expressive approximation, not an imitation. It preserves the \textbf{persona and rhythm} of the character while avoiding the reproduction of identifiable voiceprints. This approach minimizes ethical and legal concerns while enabling consistent, role-aware voice synthesis across narrative scenes.

\subsection{Additional Analyses: Voice Consistency, Persona Modeling, and Robustness}
\label{supplementary:additional_analyses}

\subsubsection{Voice Consistency Across Characters.}
To further address reviewer concerns regarding occasional inconsistencies in generated speech, particularly for characters with monotonic delivery and low lexical diversity, we present a quantitative analysis of the reference voice prompts. Table~\ref{tab:voice-stats} summarizes key lexical and acoustic properties across a diverse set of character types. 

\begin{table}[h]
\centering
\small
\begin{tabular}{lrrrrr}
\toprule
\textbf{Character} & \textbf{Unique Words} & \textbf{Voiced Phonemes} & \textbf{Pitch Std (Hz)} & \textbf{Pause Ratio} & \textbf{Duration (sec)} \\
\midrule
Shrek     & 23 & 70  & 57.77  & 1.00 & 10.00 \\
Donkey    & 38 & 118 & 83.79  & 1.00 & 10.00 \\
Doraemon  & 23 & 70  & 106.73 & 0.98 & 8.37  \\
Nobita    & 24 & 59  & 146.51 & 0.94 & 9.81  \\
Kevin     &  9 & 33  & 79.80  & 1.00 & 5.24  \\
Bob       & 11 & 40  & 139.87 & 1.00 & 9.47  \\
Tom       & 28 & 73  & 127.81 & 1.00 & 12.68 \\
Jerry     & 18 & 57  & 136.64 & 1.00 & 8.16  \\
\bottomrule
\end{tabular}
\caption{\textbf{Reference Prompt Statistics.} Voice prompts differ significantly in lexical richness, prosodic dynamics, and temporal structure. High pitch variability does not necessarily correlate with expressive or consistent speech style; utterances with low lexical variety and near-continuous voicing (pause ratio $\approx$ 1.0) tend to sound more monotonic. This suggests that consistent speech perception depends on the interplay between lexical diversity, phoneme variation, and rhythmic structure—not pitch alone.}
\label{tab:voice-stats}
\end{table}

\subsubsection{Voice Consistency Across Characters: Metric Descriptions.}
Each column in Table~\ref{tab:voice-stats} is computed to quantify the linguistic and prosodic variability of the reference voice prompts:

\begin{itemize}
    \item \textbf{Unique Words:} The number of distinct alphabetic tokens in the transcript, obtained using the NLTK tokenizer. This reflects lexical diversity, which contributes to the perceived richness and individuality of a character's speech.

    \item \textbf{Voiced Phonemes:} The number of voiced phonemes (e.g., /b/, /m/, /a/) in the transcript, extracted using the \texttt{phonemizer} library with the espeak backend (en-us). This metric captures phonetic variety and articulation complexity.

    \item \textbf{Pitch Std (Hz):} The standard deviation of the estimated fundamental frequency (F0), computed using WORLD vocoder methods (\texttt{pyworld.harvest} and \texttt{stonemask}). Higher values indicate greater prosodic variation and expressive tone.

    \item \textbf{Pause Ratio:} The proportion of silent frames in the audio, based on short-time energy computed with a sliding window (frame length = 2048, threshold = 0.01). A ratio near 1.0 indicates continuous voicing with few pauses, often perceived as monotonic delivery.

    \item \textbf{Duration (sec):} Total length of the prompt audio, calculated as the number of samples divided by the sampling rate.
\end{itemize}

These metrics jointly characterize the structure and expressivity of each prompt. We observe that characters with high pitch variance but low lexical diversity and minimal pauses (pause ratio $\approx 1.0$) tend to exhibit more frequent identity drift over long conversations. This suggests that speaker consistency is influenced not just by acoustic features, but also by linguistic variety and temporal structure. Future work may explore multi-turn prosody conditioning or speaker-aware memory to improve stability for such monotonic speech patterns.

\subsubsection{Character Persona Modeling.}
Our system does not rely on static personality templates or trainable embeddings. Instead, character traits are dynamically inferred from the interplay of:
\begin{itemize}
\item the scene and action prompts,
\item keyframe image captions (BLIP),
\item and recursively accumulated dialogue history (via RNB).
\end{itemize}

This flexible mechanism allows for emergent behaviors and contextually appropriate responses, even for unseen or sparsely referenced characters like Tom and Jerry—who, as you may recall, rarely speak at all. In fact, their only widely known speaking moment comes from the delightfully controversial 1992 film Tom and Jerry: The Movie, which we bravely use as our sole reference. Rather than a limitation, we consider this an extreme zero-shot challenge: can the system generate character-faithful dialogue for two icons of silence? While the movie may have divided fans, our pipeline passed the test—producing speech that’s chaotic, emotionally erratic, and somehow still on-brand. Just like Tom and Jerry themselves.

\subsubsection{Memory Length and Long-form Generation.}
The Recursive Narrative Bank (RNB) retains \textit{all} prior utterances by default (N = all), ensuring long-range dependency modeling and character continuity. Each dialogue turn remains compact (less than 100 characters), resulting in manageable prompt lengths. We have successfully generated stories with over 13 prior utterances without encountering token limits or degradation. This design supports scalable storytelling over extended sequences without truncation or loss of coherence.

\subsubsection{Robustness to Model Substitution.}
Our modular pipeline decouples each stage (video, dialogue, speech) using intermediate natural-language representations. When we substitute the visual module (e.g., replacing Text2Story with Mochi or Vlogger), downstream components remain stable. As illustrated in Figure~\ref{fig:ablation_study}, quality degrades only locally (e.g., visual-semantic alignment) without cascading failures. This architecture supports component-level improvements and domain transfer without retraining the full pipeline.

\end{document}